\documentclass[10pt,twocolumn,letterpaper]{article}

 \usepackage[pagenumbers]{cvpr} 

\usepackage{todonotes}
\usepackage{pifont}
\usepackage{tikz}
\usepackage{adjustbox}
\usetikzlibrary{shapes,arrows,positioning}

\usepackage[acronym]{glossaries}
\newacronym{ALS}{ALS}{airborne laser scanning}
\newacronym{MLS}{MLS}{mobile laser scanning}
\newacronym{LoD}{LoD}{level of detail}
\newacronym{LoDs}{LoDs}{level of details}
\newacronym{OGC}{OGC}{Open Geospatial Consortium}
\newacronym{GML}{GML}{Geography Markup Language}
\newacronym{ASAM}{ASAM}{Association for Standardization of Automation and Measuring Systems}
\newacronym{TLS}{TLS}{terrestrial laser scanning}
\newacronym{UAV}{UAV}{unmanned aerial vehicle}
\newacronym{HD}{HD}{high definition}
\newacronym{RANSAC}{RANSAC}{RANdom SAmple Consensus}
\newacronym{ROI}{ROI}{region of interest}
\newacronym{DEM}{DEM}{digital elevation model}
\newacronym{ICP}{ICP}{iterative closest point}
\newacronym{NLOS}{NLOS}{non-line-of-sight}
\newacronym{SfM}{SfM}{structure from motion}
\newacronym{FME}{FME}{Feature Manipulation Engine}
\newacronym{OSM}{OSM}{OpenStreetMap} 
\newacronym{RMSE}{RMSE}{root mean square error}
\newacronym{CPT}{CPT}{conditional probability table}
\newacronym{DST}{DST}{Dempster–Shafer theory}
\newacronym{BN}{BayNet}{Bayesian network}
\newacronym{GIS}{GIS}{Geographic Information System}
\newacronym{PPD}{PPD}{posterior probability distribution}
\newacronym{CI}{CI}{confidence interval}
\newacronym{IFC}{IFC}{Industry Foundation Classes}
\newacronym{CRS}{CRS}{coordinate reference system}
\newacronym{LoFG}{LoFG}{Level of Facade Generalization}
\newacronym{GSV}{GSV}{Google Street View}
\newacronym{PCA}{PCA}{principal component analysis}
\definecolor{cvprblue}{rgb}{0.21,0.49,0.74}
\usepackage[pagebackref,breaklinks,colorlinks,allcolors=cvprblue]{hyperref}

\def\paperID{4} 
\def\confName{CVPR}
\def\confYear{2025}

\title{Texture2LoD3: Enabling LoD3 Building Reconstruction With Panoramic Images} 

\author{Wenzhao Tang\textsuperscript{*1 }, Weihang Li\textsuperscript{*1,2 }, Xiucheng Liang\textsuperscript{3 }, Olaf Wysocki\textsuperscript{1 }, \\ Filip Biljecki\textsuperscript{3 }, Christoph Holst\textsuperscript{1 }, Boris Jutzi\textsuperscript{1 } \\ \\
\textsuperscript{1 }Technical University of Munich, \textsuperscript{2 }Munich Center for Machine Learning, \\\textsuperscript{3 }National University of Singapore\\
{\tt\small (wenzhao.tang, ... boris.jutzi)@tum.de} ;
{\tt\small (xiucheng, filip)@nus.edu.sg}\\
{\tt\small * equal contribution}\\
}

\begin{document}
\maketitle
\begin{abstract}
Despite recent advancements in surface reconstruction, Level of Detail (LoD) 3 building reconstruction remains an unresolved challenge.
The main issue pertains to the object-oriented modelling paradigm, which requires georeferencing, watertight geometry, facade semantics, and low-poly representation -- Contrasting unstructured mesh-oriented models. 
In Texture2LoD3, we introduce a novel method leveraging the ubiquity of 3D building model priors and panoramic street-level images, enabling the reconstruction of LoD3 building models.
We observe that prior low-detail building models can serve as valid planar targets for ortho-rectifying street-level panoramic images. 
Moreover, deploying segmentation on accurately textured low-level building surfaces supports maintaining essential georeferencing, watertight geometry, and low-poly representation for LoD3 reconstruction.
In the absence of LoD3 validation data, we additionally introduce the ReLoD3 dataset, on which we experimentally demonstrate that our method leads to improved facade segmentation accuracy by 11\% and can replace costly manual projections.
We believe that Texture2LoD3 can scale the adoption of LoD3 models, opening applications in estimating building solar potential or enhancing autonomous driving simulations.
The project website, code, and data are available here: \href{https://wenzhaotang.github.io/Texture2LoD3/}{https://wenzhaotang.github.io/Texture2LoD3/}.

\end{abstract}    
\section{Introduction}
\label{sec:intro}
\begin{figure}[t]
    \centering
    \includegraphics[width=0.8\linewidth]{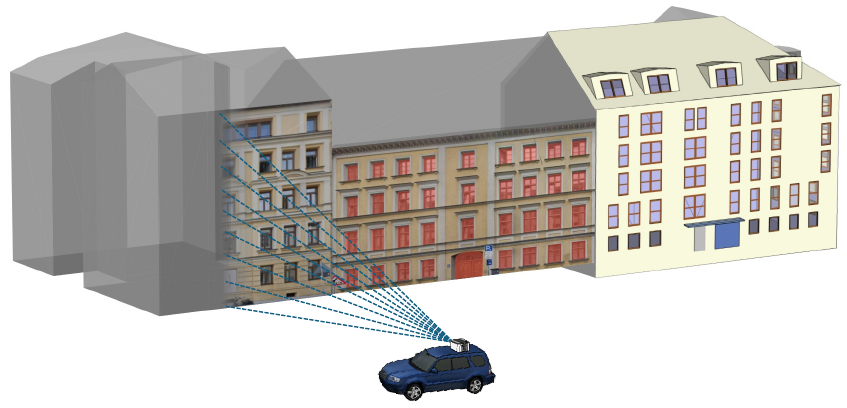}
    \caption{Texture2LoD3 proposes leveraging ubiquitous street-level images and low-level building models for accurate ortho-texturing (left): Enabling accurate semantic segmentation (center) and facade-rich \gls{LoD}3 reconstruction (right).}
    \label{fig:teaser} 
\end{figure}
Photogrammetry and computer vision researchers have always seen detailed semantic 3D building reconstruction as a fundamental challenge~\cite{szeliski2010computer,HAALA2010570}.
Recent developments in open source and proprietary software have shown that reconstruction using 2D building footprints and aerial observations enables country-wide reconstruction up to the \gls{LoD}2 displaying complex roof shapes and simplified facades~\cite{RoschlaubBatscheider,HAALA2010570,wysocki2024reviewing}.
Unlike the mesh-oriented models, the semantic 3D building models defined by the international CityGML standard \cite{grilli2019geometric} are georeferenced, watertight, and have low-poly boundary representation (B-Rep), enabling multiple applications \cite{biljeckiApplications3DCity2015}. 
Remarkably, such models remain under-explored modality for methods development, given their ubiquity, e.g., open data on 215 million buildings in Switzerland, the Netherlands, the US, or Poland \cite{wysocki2024reviewing,awesomeCityGML}.

Unlike low-detail \gls{LoD}1 and \gls{LoD}2, \gls{LoD}3 models are characterized by additional detailed facade representation and remain scarcely available despite novel methods presence \cite{wysocki2023scan2lod3,pantoja2022generating,helmutMayerLoD3,hoegner2022automatic,wang2024framework}.
One of the main issues pertains to the source data availability, assuming either accurate \gls{MLS} observations or ortho-rectified textures, which in practice are often unavailable.

Despite worldwide availability of panoramic street-level images such as \gls{GSV} or Mapillary \cite{hou2024global} and the growth in image-based training datasets, facade elements remain frequently unlabeled and limited to ortho-rectified image views \cite{wang2024framework,wysocki2023scan2lod3,Tylecek13,korc2009etrims}.
Applying such training sets to perspective and panoramic images remains unfeasible due to drastic geometry representation changes in facade elements, e.g., the closer rectangular windows are to the vanishing point, the more they resemble lines.

As we exemplify in \cref{fig:teaser}, our Texture2LoD3 proposes a method harnessing the potential of widely available panoramic street-view images and ubiquitous low-level semantic 3D building models.
We leverage the georeferencing of two modalities for their global matching while low-poly planar representation of 3D models for the image ortho-rectification target.
By utilizing prior low-poly models, we satisfy requirements of georeferencing, watertightness, low-poly representation, and geometrical consistency for \gls{LoD}3 reconstruction: Formulating it as a refinement strategy \cite{wysocki2023scan2lod3} of low-level models to high-detail \gls{LoD}3 models by reconstructing only the required facade elements, segmented from a projected image onto a planar surface.
%
%
Our main contributions are as follows:

\begin{itemize}
    \item We propose the effective projection of panoramic images to ortho-rectified images by leveraging ubiquitous semantic 3D building models as targets
    \item We improve facade semantic segmentation performance on 3D surfaces by accurate texturing: Enabling accurate \gls{LoD}3 facade element reconstruction
    \item We introduce the first-of-its-kind open texturing benchmark dataset, ReLoD3, comprising synchronised \gls{LoD}3 models, panoramic images, and manually textured low-level \gls{LoD}2 building models 
\end{itemize}
\section{Related Works}
\label{sec:rw}

%

\noindent \textbf{3D Facade Segmentation}
The recent years have witnessed a surge in semantic 3D facade segmentation methods both on point clouds, images, and in combination with prior 3D models.
Since the current research suggests that street-level and drone-based point clouds accurately depict 3D facade geometry, multiple point-cloud-based methods have been proposed \cite{pierdicca2020point,matrone2020comparing,yuetanDeepLearningOfficial,grilli2020machine}.
Recent benchmark data results, such as ZAHA \cite{Wysocki_2025_WACV} and ArCH \cite{matrone2020comparing}, imply that the challenge is still unsolved and remains challenging due to under-represented classes, sparsity of objects in point clouds, and frequently indistinct 3D geometry features \cite{su2022dla, romanengo2025discretisation}.

Other approaches rely only on image-based input, capitalizing on rich optical features and 2D image grid representation.
Various methods have been proposed to tackle this challenge, such as non-learning \cite{szeliski2010computer,musialski2013survey}, gramma-based \cite{MAYERejMCMC,brenner2006extraction}, and recently deep learning approaches \cite{liu2020deepfacade,KadaFacades,helmutMayerLoD3,wang2024framework,ceron2025oa}.
Owing to the ubiquity of image training data, even the standard Mask-RCNN \cite{he2017mask} proves relatively efficient after the subsequent fine-tunning on the facade image databases \cite{wysocki2023scan2lod3}.
However, these methods perform well only under the assumption that an image is ortho-rectified; it makes generalization challenging since facade elements are prone to the dire geometry change under perspective and barrel distortions.
This applies to classical methods as well which explicitly concentrate on line and point extraction for matching images with models for texturing \cite{kada2005facade,tan2008large,iwaszczuk2011matching,hoegner2007texture}.
In practice, ortho-rectified images are rare and limited just to a few benchmarks or manual projections, yielding unsatisfactory results on non-rectified real-world data \cite{3dgeoinfo2024Texturing,Tylecek13,korc2009etrims,gadde2016learning,riemenschneider2012irregular,kelly2024winsyn}.

An alternative approach is to exploit information from 3D models, optical images, and laser scanning point clouds to achieve accurate 3D facade segmentation \cite{tuttas_reconstruction_2013,wysocki2023scan2lod3}.
For example, Scan2LoD3 \cite{wysocki2023scan2lod3} introduces a method where uncertainty-aware ray analysis of laser points with 3D models yield conflict maps indicating openings, which can serve as evidence for late-fusion of 3D segmented point clouds and 2D segmented optical images. 
However, the availability of such multi-modal setups is currently limited and assumes their heterogeneous accurate projection onto the model surface.

\noindent \textbf{LoD3 Building Reconstruction}
Semantic 3D building reconstruction is a long-standing challenge in photogrammetry and computer vision \cite{szeliski2010computer}.
For years, the international standard CityGML \cite{grogerOGCCityGeography2012,biljecki2014formalisation} has been defining the formal description of such models, where \gls{LoD}1 stands for simple cuboid models, \gls{LoD}2 for polyhedral models with detailed roof shape, and \gls{LoD}3 for detailed roof shapes complemented with a detailed facade representation.
The primary difference to the standard mesh models is that semantic 3D building models are georeferenced; comprise object-level geometry and semantics; have a hierarchical data model that also describes the object-to-object relationship; display watertight and low-poly geometry facilitating volumetric space interpretation by integrating externally observable surfaces within a boundary representation (B-Rep) \cite{Kolbe2021,grogerOGCCityGeography2012,wysocki2024reviewing}.
%

Despite recent advancements in \gls{LoD}3 building reconstruction, \gls{LoD}3 models remain scarce~\cite{wysocki2023scan2lod3, pantoja2022generating, pang20223d, KadaFacades, wang2024framework, harshit2024low,salehitangrizi20243d}. 
One of the main remaining issues is the robustness of methods when deployed at scale.
\begin{figure*}
    \centering
    \includegraphics[width=0.95\linewidth]{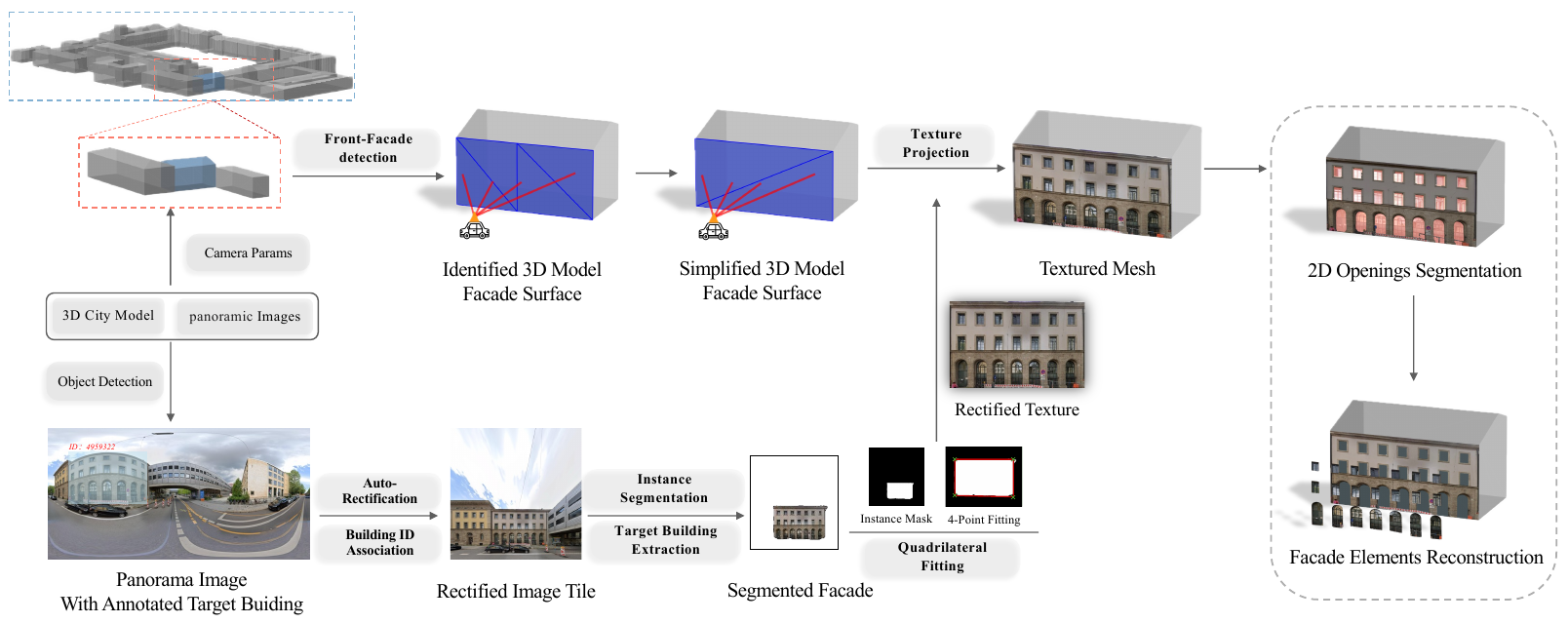}
    \caption{Overview of the proposed Texture2LoD3 method: The method commences with global matching of georeferenced panorama images and low-level 3D models. In the top branch, 3D target facade surfaces are simplified, while in the bottom branch panoramic images are rectified and building facade instances are extracted. Subsequently, fine object-to-object matching and projection is performed to the simplified 3D model surface. Quadrilateral fitting and image-to-plane ray casting ensure accurate ortho-rectified 3D texture, enabling accurate facade elements segmentation and \gls{LoD}3 reconstruction.  }
    \label{fig:overview}
\end{figure*}
Most of the methods assume that a specialized method of acquisition is required: 
It sets a high requirement for the practical methods' deployment, as these methods assume targeted accurate co-registration of multiple subsequent images and complete object coverage without adjacent buildings, e.g., single house acquired by a 360-degree drone flight \cite{pantoja2022generating,helmutMayerLoD3}.
%
%
Alternatively, the above-mentioned Scan2LoD3 \cite{wysocki2023scan2lod3} can mitigate such issues by introducing additional conflict maps of the ray-to-prior-model analysis.
%
%

\section{Method}
\label{sec:method}
As shown in~\cref{fig:overview}, our Texture2LoD3 method commences with the image-to-object matching of widely-available geo-referenced panoramic images and ubiquitous low-level semantic 3D building models (\cref{sec:wireframe_preprocessing}).
This process is followed by 3D model B-Rep surface simplification (top-branch), while panoramic images are rectified (\cref{sec:panoramic_rectification}) and building facades are segmented (bottom-branch) (\cref{sec:facade_seg}).
The fine quadrilateral fitting of the facade instance shall ensure complete facade depiction (\cref{sec:morph_adapt}), followed by ray-casting-based projection onto the simplified 3D model planar surface (\cref{sec:texturing}); Enabling accurate facade elements segmentation and \gls{LoD}3 reconstruction.

\subsection{Matching Panoramic Image to 3D Model}
\label{sec:wireframe_preprocessing}

In this work, “matching” refers to aligning geo-referenced ground-level panoramic images with corresponding 3D building models. Specifically, the goal is to associate the facade observed in a ground-level image with its counterpart in the 3D model, thus establishing a coherent mapping between image pixels and 3D geometry.

\noindent \textbf{Camera Parameters}  
We assume that each panoramic image is accompanied by a set of camera parameters that are essential for the matching process. In particular, the camera parameters include:
a) \textit{Position}: The geographic coordinates (latitude and longitude) of the camera; b) \textit{Heading}: The azimuth angle indicating the direction the camera faces, measured in degrees clockwise from North; c)  \textit{Field-of-view (FOV)}: The angular extent of the scene captured by the camera in degrees; d) \textit{Generic parameters}: Any extra available parameters, e.g., the camera's height above ground level.

Due to the imprecision of geo-referenced data, the available 2D sensor positions and 3D model vertices in the B-Rep only provide a coarse association. Moreover, semantic 3D building models often subdivide a single facade into multiple small triangular faces—a phenomenon we refer to as facade subdivision. This subdivision complicates texture mapping because it prevents a straightforward correspondence between image features and continuous facade regions. To overcome these issues, we propose a unified ray-casting-based approach that leverages camera parameters to detect facade regions and simplify the 3D model, thereby facilitating a robust matching between the panoramic image and the 3D building model.


\noindent \textbf{3D B-Rep Model and Camera Integration} 
We first extract the camera parameters (position, heading, FOV, and the manually set camera height) from the geo-referenced panoramic images and project them into the global building coordinate reference system. In our framework, the 3D building model is represented as a boundary representation (B-Rep), i.e., a collection of vertices, edges, and faces that define the surfaces of the building; 
We assume the following information is available:
a) \textit{Ground surface definition}: The model explicitly delineates the building's base, from which the building height can be extracted, e.g., via the minimum and maximum Z-coordinates adhering to the CityGML GroundSurface definition \cite{grogerOGCCityGeography2012};
b) \textit{Altitude and orientation}: The model's global orientation (altitude) is inherently defined within a global coordinate reference system, ensuring that facade orientations are consistent;
c) \textit{Height}: The vertical extent of the building is provided or can be computed from the B-Rep, enabling precise placement of the camera.

\noindent \textbf{Ray-Casting-Based Facade Detection} 
For each camera, multiple rays with varying horizontal and vertical angles are cast against the 3D model's triangular mesh. 
The ray-casting process records the intersected faces and their spatial distribution. 
We then select the camera view that yields the highest number of valid intersections and best aligns the camera position with the centroid of the hit points. 
Note that here, ray-casting is used to robustly detect the facade region by identifying the contiguous set of faces corresponding to the building’s facade, even in the presence of fragmentation. 
This detection step is crucial for the subsequent matching process, as it determines which part of the 3D model corresponds to the observed image.

\begin{figure}[t]
    \centering
    \includegraphics[width=0.5\linewidth]{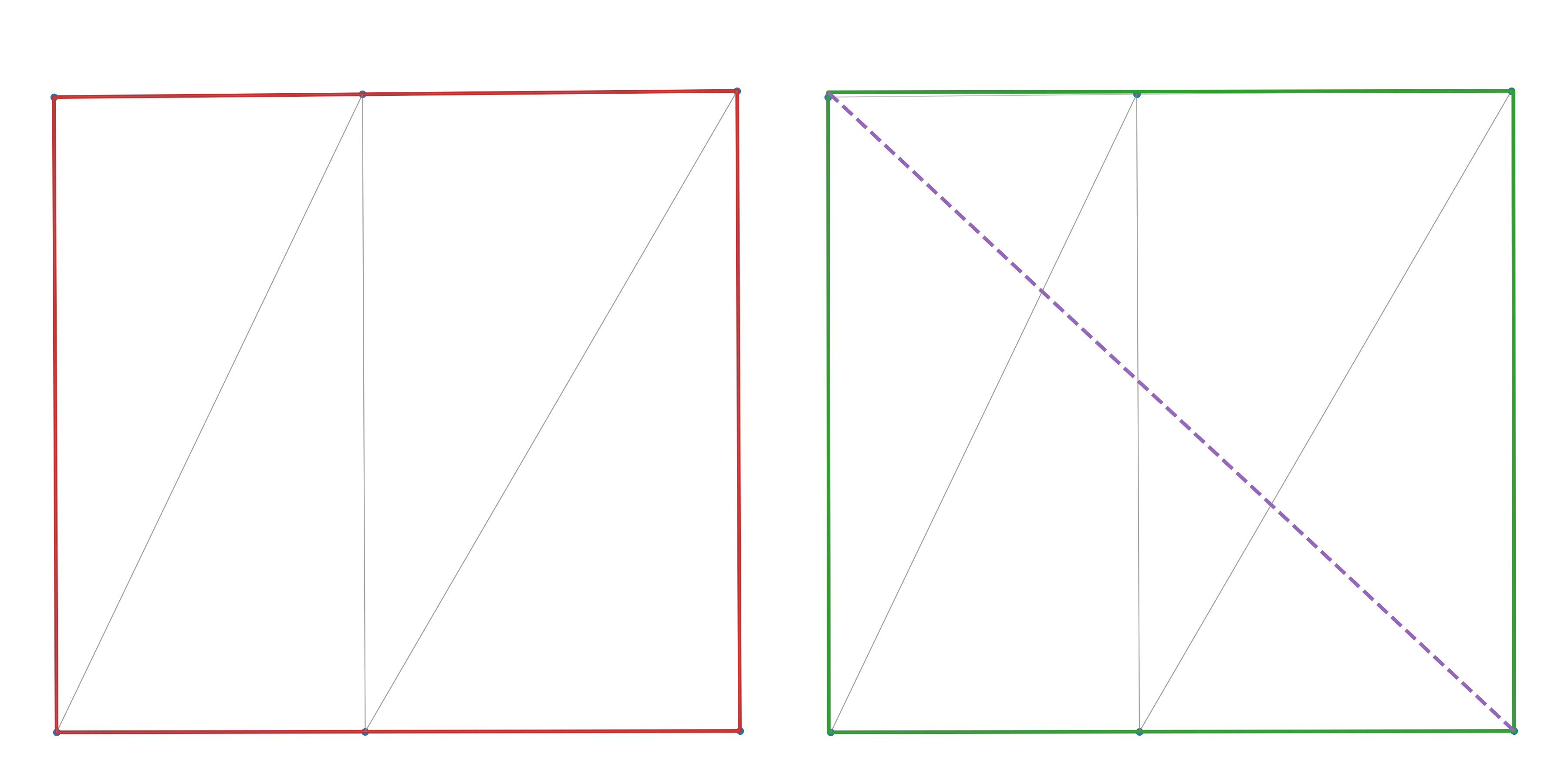}
    \caption{(Left) Original surface with multiple triangular faces. 
    (Right) Fitted quadrilateral representation with re-triangulation along the diagonal (dashed purple), preserving facade shape. }
    \label{fig:wireframe_comparison}
\end{figure}

\noindent \textbf{Local Plane Fitting and 3D B-Rep Model Simplification} 
The set of intersected triangular faces from the optimal view is aggregated and fitted to a local plane via \gls{PCA}, which yields a centroid \(\mathbf{c}\) and two in-plane basis vectors \(\mathbf{u}\) and \(\mathbf{v}\). Each vertex \(p\) on the detected facade is then projected onto this plane:
\begin{equation}
x = \langle p - \mathbf{c}, \mathbf{u} \rangle, \quad y = \langle p - \mathbf{c}, \mathbf{v} \rangle
\end{equation}
From the 2D projections, a minimum area bounding rectangle is computed, resulting in four corner points \(\{(x_i, y_i)\}_{i=1}^4\). These corners are mapped back into 3D space:
\begin{equation}
\mathbf{q}_i = \mathbf{c} + x_i\,\mathbf{u} + y_i\,\mathbf{v}, \quad i=1,\dots,4
\end{equation}
The quadrilateral defined by \(\{\mathbf{q}_i\}\) is subsequently re-triangulated into two triangles, thereby replacing the fragmented original representation with a simplified mesh that preserves critical geometric features while reducing computational complexity (\cref{fig:wireframe_comparison}). 
%

It is worth noting that the literature offers a wide variety of methods for geometric simplification and for converting between triangular and quadrilateral representations, such as those available in the CGAL library \cite{fabri2009cgal}.
In contrast, our approach relies solely on the consistency of plane normals computed via \gls{PCA}, which is robust under the assumption that the facade region is locally planar—a reasonable assumption for most urban building facades and semantic 3D city models that shall adhere to this assumption.

\noindent \textbf{FOV Calculation Based on Building Geometry} 
To compute the effective camera field-of-view (FOV) for each building, we define a buffer region around each camera observation point to identify nearby structures. The exterior boundaries of the building are sampled to determine the angular directions (bearings) from the camera. By evaluating occlusion effects—ensuring that each building vertex is visible without interference from adjacent structures—we determine the effective angular extent of the building facade. These FOV metrics aim to exclude occlusion-induced noise and the target building’s facade view.
%


\subsection{Panoramic Image Auto-rectification}
\label{sec:panoramic_rectification}
We utilize an automatic rectification approach for panoramic images consisting of three stages inspired by \citet{zhu2020large}: a) tile extraction and local rectification; b) consensus estimation of zenith and horizontal vanishing points; and c) global re-projection.
This part of the method aims to effectively rectify panoramic images by combining local tile analysis, a robust SVD-based consensus, and global re-projection.
It shall provide a consistent geometric basis for subsequent facade segmentation and texturing.

\noindent \textbf{Tile Extraction and Local Rectification}
We partition the input panorama image into multiple overlapping tiles via a ray-casting strategy. Local features and edges within each tile yield estimates of the horizon line \(\mathbf{h}\), horizontal vanishing points \(\{\mathbf{v}_i\}\), and a local zenith vector \(\mathbf{z}\). Importantly, the local zenith vector \(\mathbf{z}\) is computed independently from the horizontal vanishing points. Specifically, while both are derived from the same set of local edge features, the zenith vector is estimated via a robust SVD-based process on the normalized edge directions, which directly captures the predominant vertical direction in each tile. These local parameters serve as geometric cues for subsequent global alignment. Although the image is already rectified, semantic information does not drive the rectification process. Consequently, when a building's facade is particularly wide, individual tiles may only capture a portion of the facade (even if that portion is rectified). In such cases, subsequent image tile stitching (Section~\ref{sec:image_tile_stitching}) is necessary to produce a more complete representation of the facade.

\noindent \textbf{Consensus Estimation}
We aggregate all normalized zenith vectors $\{\mathbf{z}_i\}$ and compute a consensus zenith $\mathbf{z}^*$ via SVD:
\begin{equation}
   \mathbf{z}^* = \operatorname{SVD}(\{\mathbf{z}_i\}) 
\end{equation}
From $\mathbf{z}^* = (z_x, z_y, z_z)^\top$, the pitch $\phi$ and roll $\theta$ angles are:
\begin{equation}
\phi = \arctan\!\Bigl(\frac{z_z}{z_y}\Bigr), 
\quad 
\theta = -\arctan\!\Bigl(\frac{z_x}{\operatorname{sgn}(z_y)\sqrt{z_y^2+z_z^2}}\Bigr)
\end{equation}
We define standard rotation matrices $R_{\text{roll}}(\theta)$, $R_{\text{pitch}}(\phi)$ (and optionally $R_{\text{heading}}(\psi)$) to align the vanishing points. A histogram of horizontal angles can further refine these estimates if necessary.

\noindent \textbf{Global Re-projection}
With the consensus rotation determined, we re-project the entire panorama image into a rectified view. For a pixel with spherical coordinates $(\theta, \phi)$, its 3D direction vector $\mathbf{v}(\theta, \phi)$ is rotated back by $R_{\text{roll}}(-\theta)$ and $R_{\text{pitch}}(-\phi)$. The result is then mapped to image coordinates via an inverse equirectangular projection:
\begin{equation}
x = \Bigl(\tfrac{\theta'}{360^\circ} + \tfrac12\Bigr)W,\quad
y = \Bigl(\tfrac{\phi'}{180^\circ} + \tfrac12\Bigr)H
\end{equation}
%

\noindent \textbf{Image Tile Stitching}
\label{sec:image_tile_stitching}
In cases where a single rectified tile cannot capture the entire building facade, 
we stitch multiple overlapping tiles into one image. 
We detect SIFT keypoints in each tile, 
match them across overlaps, and estimate a robust homography via RANSAC~\cite{linrl3_image_stitching}. 
The source tile is then warped accordingly, and a smooth blending operation mitigates seam artifacts. 
%

\subsection{Building Facade Segmentation}
\label{sec:facade_seg}
\begin{figure}[t]
    \centering
    \includegraphics[width=0.95\linewidth]{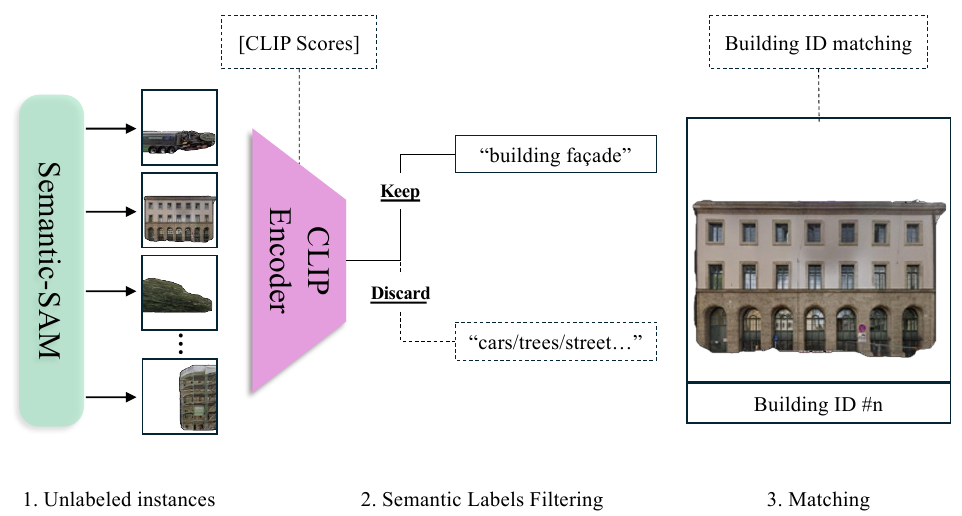}
    \caption{\normalsize Semantic-SAM generates unlabeled instance masks, which are then passed to a CLIP encoder for semantic filtering. We retain masks classified as \emph{building facade}.}
    \label{fig:facade_seg_pipeline}
\end{figure}

To accurately isolate and extract building facades from complex urban scenes, we adopt the pipeline illustrated in~\cref{fig:facade_seg_pipeline}. 
Our approach integrates an automatic instance‐level segmentation (Semantic-SAM~\cite{li2023semantic}) with semantic filtering via CLIP~\cite{radford2021learning}, thus allowing building facades to be selectively retained while discarding irrelevant objects (e.g., cars, trees, people). 
We choose Semantic-SAM owing to its outstanding performance in instance segmentation tasks~\cite{li2023semantic}. 
Given that many of our input images feature multiple adjacent building facades, Semantic-SAM’s robust segmentation capability is essential for reliably distinguishing individual facade instances.

\noindent\textbf{Instance Generation via Semantic-SAM}  
Given a rectified panoramic image $I$, we employ the Semantic-SAM automatic mask generator to produce a set of unlabeled instance masks $\{M_i\}$. These masks aim to cover all salient regions in the scene, ranging from building surfaces to smaller objects like cars or trees. Although the mask generator provides instance-level segmentation, no semantic labels are assigned.

\noindent\textbf{CLIP-Based Label Filtering}
To determine which instance masks correspond to building facades, we process each masked image region using a CLIP~\cite{radford2021learning} encoder (ViT-L/14). Specifically, we compute an image embedding and compare it via cosine similarity to text embeddings derived from a predefined set of text prompts (typically 2–3 prompts, e.g., \emph{"building facade"}, \emph{"vehicle"}, and \emph{"pedestrian on the street"}).
An instance is retained if its highest-confidence label is \emph{"building facade"} and its similarity score exceeds a chosen threshold; otherwise, it is discarded. Additionally, masks identified as \emph{"building eave"} are subtracted to ensure that only the primary vertical surfaces of the building remain. This process also filters out instances classified as \emph{"vehicle"} or \emph{"pedestrian on the street"} to exclude dynamic and non-architectural elements from further processing.

\noindent\textbf{Mask Combination and Noise Removal}  
As multiple facade masks may be produced for a single building or portions thereof, we unify them via logical OR:
\( M_{\text{facade}} = \bigvee_{i \in \mathcal{I}} M_i \), where each \( M_i \) is "building facade".
Likewise, all eave masks are aggregated via logical OR and then subtracted from $M_{\text{facade}}$. We further remove small connected components whose area is below a minimum threshold, estimated by $A_{\min}$, to eliminate spurious detections. 
Morphological opening and closing are then performed using a kernel of size $k\times k$ (with $k$ chosen according to the image resolution) to fill small holes and smooth the boundaries of the combined mask.
%

\noindent \textbf{Final Facade Extraction}  
After noise removal, the resulting binary mask accurately outlines the dominant building facades. 
As a final step, we align the mask size with the original panorama image and multiply it element‐wise with the original image \( I_{\text{masked}}(x,y) = I(x,y) \times M_{\text{facade}}(x,y) \),
yielding a facade‐only color image that is preserved for subsequent morphological adaptation (\cref{sec:morph_adapt}) and texturing (\cref{sec:texturing}).
\begin{figure}[t]
    \centering
    \includegraphics[width=0.8\linewidth]{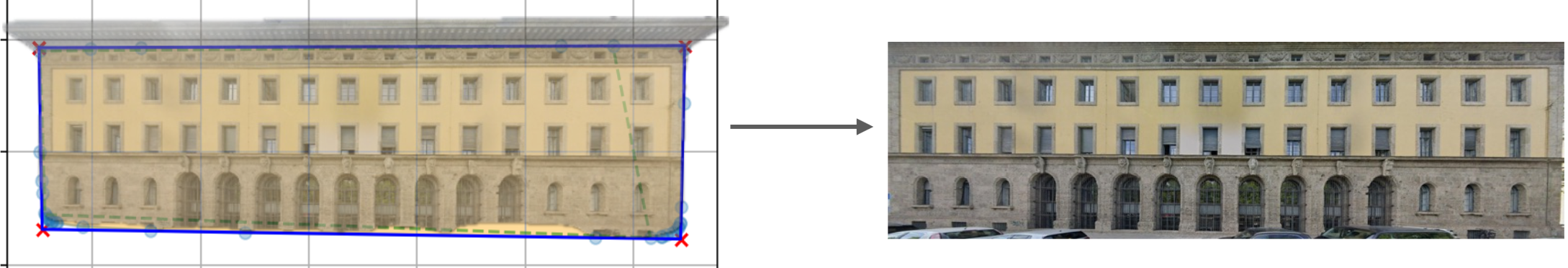}
    \caption{Building facade after filtering out extraneous parts, e.g., eave masks, and a schematic view  quadrilateral fitting extracting the four corner points based on the refined mask.}
    \label{fig:quadrilateral_fitting_demo2}
\end{figure}



\subsection{Facade Mask Quadrilateral Fitting}
\label{sec:morph_adapt}
In this step, we refine the facade segmentation mask to produce a clean, noise-free representation that accurately outlines the facade (\cref{fig:quadrilateral_fitting_demo2}). The process consists of three stages: a) mask smoothing via morphological operations; b) robust quadrilateral fitting to the facade contour; and c) perspective rectification.

\noindent \textbf{Smoothing via Morphological Operations}  
Given an input binary mask $I$, we first smooth the mask by applying a Gaussian blur:
\begin{equation}
I_{\text{blur}}(x,y) = \sum_{(u,v) \in \Omega} G(u,v,\sigma) \, I(x-u, y-v),
\end{equation}
where $G(u,v,\sigma)$ is a Gaussian kernel and $\Omega$ is the kernel support. This smoothing reduces high-frequency noise. Next, we perform morphological closing followed by opening to fill small holes and remove spurious regions:
\begin{equation}
I_{\text{close}} = (I_{\text{blur}} \oplus B) \ominus B, \quad
I_{\text{open}} = (I_{\text{close}} \ominus B) \oplus B,
\end{equation}
with $\oplus$ and $\ominus$ denoting dilation and erosion, respectively, and $B$ being a rectangular structuring element of size $(15\times15)$. From the resulting mask, contours are extracted and the largest contour, $C_{\max}$, is selected:
\( C_{\max} = \arg\max_{C \in \mathcal{C}} \; \text{Area}(C) \).
Its convex hull, \( H = \operatorname{convexHull}(C_{\max}) \),
provides a robust boundary for the facade.

\noindent \textbf{Quadrilateral Fitting}  
To obtain a compact facade representation, we fit a quadrilateral to the points of the convex hull. Let $\mathcal{P} = \{p_1, p_2, \ldots, p_n\}$ denote the set of points in $H$. We seek a quadrilateral $Q$ with vertices $\{q_1, q_2, q_3, q_4\}$ that maximizes the Intersection over Union (IoU) with $H$, where:
\begin{equation}
\text{IoU}(H, Q) = \frac{\text{Area}(H \cap Q)}{\text{Area}(H \cup Q)}
\end{equation}

\noindent \textbf{Perspective Rectification}  
With the scaled quadrilateral $Q^{\text{scaled}}$, we compute a homography that maps its vertices to the corners of a target rectangle. Assuming that the target image has width $W$ and height $H$, we define:
\begin{equation}
T = \{ (0,0), \, (W-1,0), \, (W-1,H-1), \, (0,H-1) \}
\end{equation}
The homography matrix $P$ satisfies:
\begin{equation}
\begin{bmatrix} x'_i \\ y'_i \\ 1 \end{bmatrix} \sim P \begin{bmatrix} x_i \\ y_i \\ 1 \end{bmatrix}, \quad i=1,\dots,4
\end{equation}
where $(x_i,y_i)$ are the coordinates of $q_i^{\text{scaled}}$ and $(x'_i,y'_i)$ are the corresponding target coordinates. 
This perspective transformation matrix is computed and applied to the original image:
\( I_{\text{warped}} = \operatorname{warpPerspective}(I_{\text{orig}}, P) \).
%


\subsection{Facade Texturing by Ray-Casting}
\label{sec:texturing}
In this stage, we accurately map the texture from the panoramic image onto the simplified facade geometry. Our approach uses a ray-casting method that projects rays from the camera center and computes their intersections with the facade surface, thus determining the texture coordinates for each sample. 

\noindent \textbf{Ray Generation and Direction Determination}  
Using the simplified facade (\cref{sec:wireframe_preprocessing}), we generate a 3D ray for each sampling point on the target texture grid. Each pixel in the panoramic image is first associated with spherical coordinates $(\theta, \phi)$, from which its 3D direction vector is computed as:
\begin{equation}
   \mathbf{v}(\theta, \phi) = \begin{bmatrix}\cos\phi\sin\theta \\ \sin\phi \\ \cos\phi\cos\theta\end{bmatrix} 
\end{equation}
Subsequently, the direction is adjusted using the inverse of the rotation matrices derived during the panoramic image auto-rectification stage for pitch, roll, and heading (\cref{sec:panoramic_rectification}). 
This step aligns the rays with the actual orientation of the facade.

\noindent \textbf{Ray-Facade Intersection}  
Each ray, cast from the camera center \(\mathbf{o}\), is tested for intersection with the facade surface. Since the facade is approximated as a quadrilateral (typically decomposed into two triangles), the intersection point is calculated using the standard ray-plane intersection formula:
\begin{equation}
t = \frac{(\mathbf{p}_0 - \mathbf{o}) \cdot \mathbf{n}}{\mathbf{v} \cdot \mathbf{n}}
\end{equation}
where \(\mathbf{p}_0\) is an arbitrary point on the facade plane, and \(\mathbf{n}\) is the unit normal vector of the plane. The intersection point is then given by \( \mathbf{p} = \mathbf{o} + t\,\mathbf{v} \).
%

While a homography warp from rectified images could be used for texture mapping, it assumes that the facade is perfectly planar and that the rectification is flawless. 
In practice, residual geometric distortions and local deviations from planarity often persist. 
Our ray-casting method directly computes the intersection of rays with the actual 3D facade, thereby accommodating these imperfections and ensuring a more robust and accurate texture mapping. 
Moreover, a simple homography warp cannot account for non-planarities or slight misalignments due to calibration errors, which our ray-casting approach inherently corrects by leveraging the true 3D geometry.

\noindent \textbf{Texture Coordinate Mapping}  
Once the intersection point \(\mathbf{p}\) is determined, it is projected onto the local 2D coordinate system of the facade using the plane parameters obtained from PCA (centroid \(\mathbf{c}\) and in-plane basis vectors \(\mathbf{u}\) and \(\mathbf{v}\)):
\begin{equation}
\label{alg:texturing_code}
x = \langle \mathbf{p} - \mathbf{c}, \mathbf{u} \rangle, \quad y = \langle \mathbf{p} - \mathbf{c}, \mathbf{v} \rangle
\end{equation}
After normalization, the \((x,y)\) coordinates correspond directly to the texture coordinates in the original panoramic image.

\noindent \textbf{Texture Sampling and Synthesis}  
The texture coordinates are used to sample pixel values from the panoramic image, employing bilinear interpolation to ensure pixel re-projection. 
These sampled values are then mapped onto the simplified facade mesh, thereby generating a high-detail, geometrically consistent texture.

\section{Experiments}
\label{sec:experiments}
\begin{figure*}[t]
    \centering
    \includegraphics[width=0.8\textwidth]{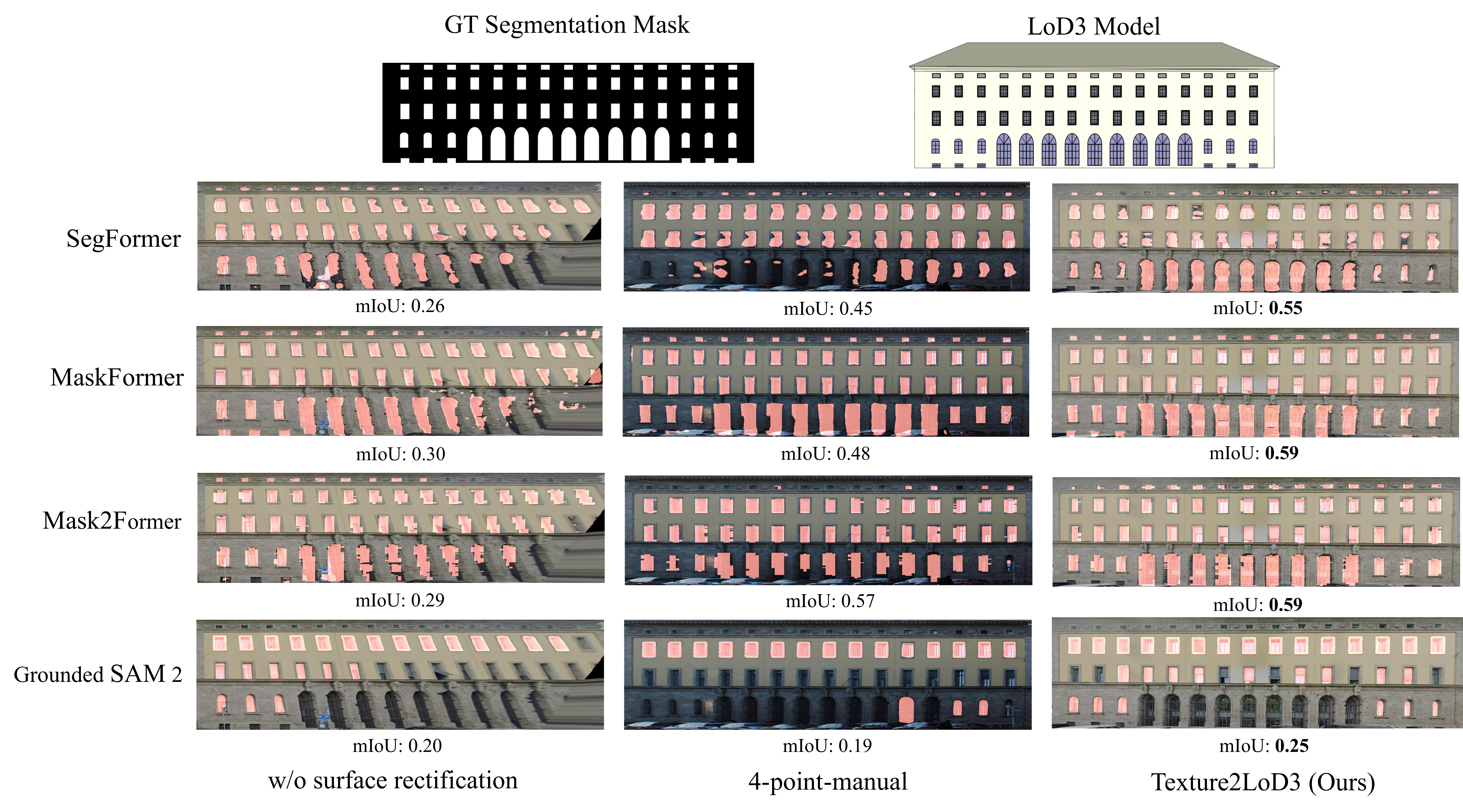}
    \caption{Tested facade segmentation baselines on a selected building from the introduced ReLoD3 benchmark dataset across various texture projection methods. Our Texture2LoD3 is less prone to distortions, hence yielding more accurate segmentation across the baselines.}
    \label{fig:segmentation_comparison_methods}
\end{figure*}
\begin{figure*}
    \centering
    \includegraphics[width=0.8\textwidth]{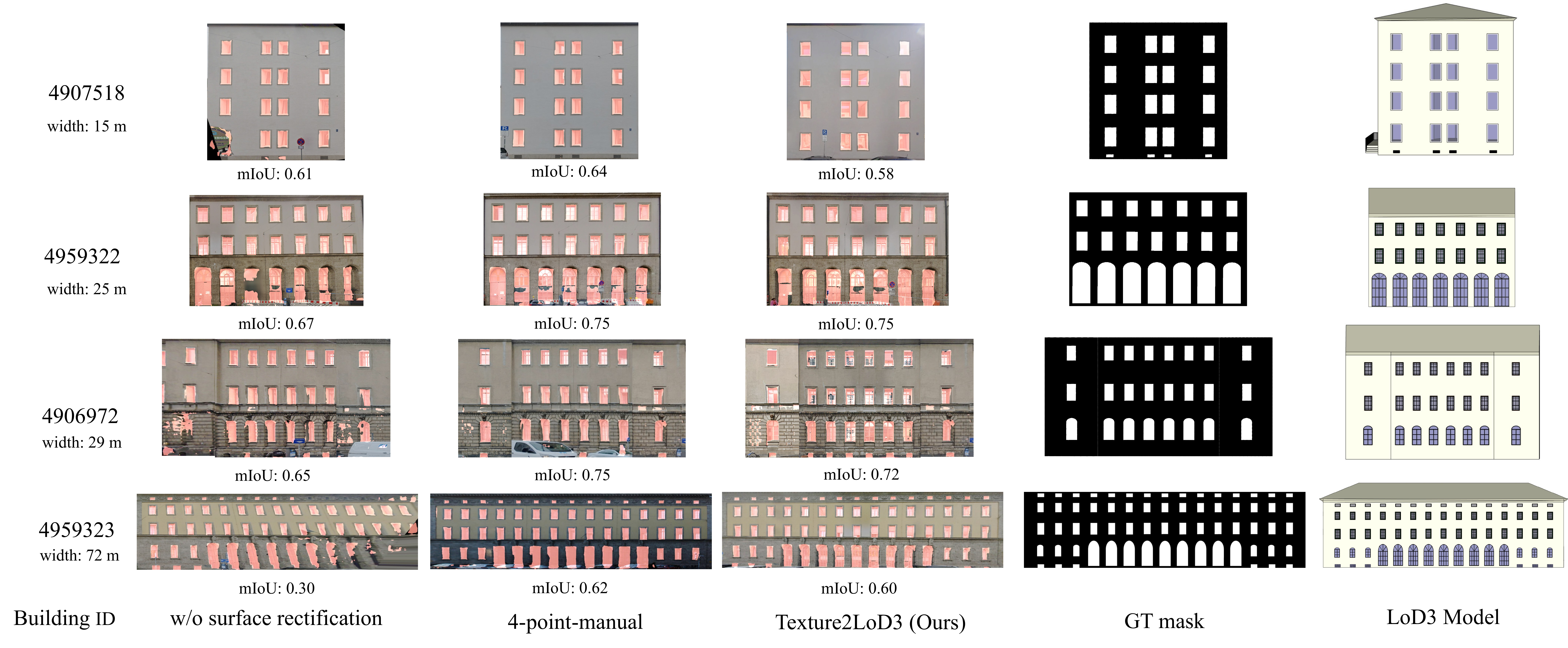}
    \caption{Texture2LoD3 maintains on-par accuracy with manual texturing even in the presence of increasing facade width, unlike the method without rectification. Shown on MaskFormer \cite{cheng2021maskformer} on four width-different facades of the introduced ReLoD3 benchmark dataset. }
    \label{fig:segmentation_comparison_building}
\end{figure*}
\begin{table}[t]
\centering
\caption{Quantitative comparison of semantic segmentation models on facade opening detection across two datasets. Performance is measured using SSIM ($\uparrow$), IoU ($\uparrow$), and LPIPS ($\downarrow$).}
\label{tab:segmentation_comparison}
\begin{adjustbox}{width=0.48\textwidth, center}
    \begin{tabular}{lccccccccc}
    \toprule
    & \multicolumn{3}{c}{w/o surface rectification 
    } & \multicolumn{3}{c}{4-point-manual} & \multicolumn{3}{c}{Texture2LoD3 (Ours)} \\
    \cmidrule(lr){2-4} \cmidrule(lr){5-7}  \cmidrule(lr){8-10}
    Methods & SSIM & IoU & LPIPS & SSIM & IoU & LPIPS & SSIM & IoU & LPIPS\\
    \midrule
    SF\cite{xie2021segformer} & 0.84 & 0.43 & 0.38 & 0.86 & 0.51 & 0.35 & \textbf{0.87} & \textbf{0.53} &\textbf{0.34} \\
    MF\cite{cheng2021maskformer} & 0.83 & 0.49 & 0.39 & \textbf{0.86} & 0.59 & 0.34 & 0.84 & \textbf{0.60} & \textbf{0.33} \\
    M2F\cite{cheng2021mask2former} & 0.84 & 0.45 & 0.37 & 0.85 & \textbf{0.48} & \textbf{0.35} & \textbf{0.86 }& \textbf{0.48} & 0.36\\
    GS2\cite{li2023semantic} & 0.83 & 0.40 & 0.39 & \textbf{0.84} & \textbf{0.44} & \textbf{0.37} & \textbf{0.84} & 0.42 & \textbf{0.37} \\
    \bottomrule
    \end{tabular}
\end{adjustbox}
\end{table}
\noindent \textbf{Our ReLoD3 Texture Dataset Benchmark} 
%
%
In the absence of datasets comprising accurate \gls{LoD}3 reference data aligned with extracted opening masks, manual textures, and street-level images, we introduce the ReLoD3 dataset.
The ReLoD3 comprises 27 unique \gls{LoD}3 models modeled according to the CityGML standard \cite{grogerOGCCityGeography2012} including windows, doors, and eaves modeled based on high-accuracy \gls{MLS} point clouds of relative accuracy 1-3 cm \cite{mofa}, manually 4-point projected perspective terrestrial optical images of the digital camera (Sony $\alpha$7), and corresponding \gls{GSV} Images \cite{googleStreetView}, located in Munich, Germany. 
This dataset is part of the TUM2TWIN initiative \cite{tum2twin}. 
We deem \gls{LoD}3 opening masks as ground-truth owing to their superior accuracy and no distortions present, unlike manually rectified perspective images.
In this experiment, we used 238 windows and 38 door instances captured from various building facades. 
The data is available under the project page: \href{https://wenzhaotang.github.io/Texture2LoD3/}{https://wenzhaotang.github.io/Texture2LoD3/}.
\subsection{Results and Discussion}
\label{sec:results}
\noindent \textbf{3D Facade Segmentation as Texture Quality Measure}
We evaluate the performance of four state-of-the-art semantic segmentation approaches on the task of facade opening detection: SegFormer \cite{xie2021segformer}, MaskFormer \cite{cheng2021maskformer}, Mask2Former \cite{cheng2021mask2former}, and Grounded SAM2 \cite{ren2024grounded} (Segment Anything Model with semantic capabilities).
For the supervised methods, we leverage the pre-trained on ADE20K \cite{zhou2017scene}, fine-tuned on the CMP dataset \cite{Tylecek13};  For the open-set experiments, we use the text prompt \textit{"window"} and \textit{"door"}. 

The quality of facade segmentation serves as an effective proxy for evaluating texture quality in 3D building models. We employ three distinct metrics to comprehensively assess segmentation performance: Structural Similarity Index Measure (SSIM) \cite{wang2004image}, mean Intersection over Union (mIoU), and Learned Perceptual Image Patch Similarity (LPIPS) \cite{zhang2018unreasonable}. SSIM measures the perceived quality between images and correlates with human visual perception, IoU quantifies the spatial overlap accuracy between predicted and ground truth segments, and LPIPS captures perceptual similarities using deep feature representations that align with human judgments of visual similarity.


We compared three texture processing methods: unrectified imagery (\textit{w/o surface rectification}), manual 4-point rectification (\textit{ReLoD3}), and our automatic approach (\textit{Texture2LoD3}). While unrectified imagery projection is the standard projection procedure, manual 4-point rectification represents the current standard in many practical workflows and relies on manual corner selection, our Texture2LoD3 uses geometric data from LoD1/2 models for automatic alignment. All images were captured at a consistent height (approx. 1.7m) to reduce alignment errors. To ensure fair segmentation comparison, we apply test-time adaptation for mask evaluation across all methods (see supplementary material for details).

As shown in \cref{tab:segmentation_comparison}, all segmentation models benefit significantly from accurate texture rectification, with consistent performance improvements visible across all metrics. The baseline approach without rectification achieves the lowest scores due to perspective distortions complicating the segmentation task. The 4-point-manual method delivers noticeable improvements, particularly in IoU scores, demonstrating the value of perspective correction in facade analysis. Our Texture2LoD3 approach consistently outperforms the w/o surface rectification baseline across all models and metrics and can replace manual projections. 
SegFormer exhibits the most substantial gains, achieving an SSIM of 0.87, IoU of 0.53, and LPIPS of 0.34 when combined with our method. This result represents improvements of 3\% in SSIM and 10\% in IoU compared to the unrectified baseline and 1-2\% improvement over the manual rectification approach.

The qualitative results in \cref{fig:segmentation_comparison_building} and \cref{fig:segmentation_comparison_methods} visually confirm these quantitative findings. \cref{fig:segmentation_comparison_building} demonstrates how our Texture2LoD3 method produces cleaner segmentation boundaries and more consistent element detection across various building facades. The improvement is particularly evident in buildings with complex architectural features and elongated facades.
%
\cref{fig:segmentation_comparison_methods} highlights performance differences through visual comparisons of a facade segmented by various methods. Texture2LoD3 produces results that align more closely with both the ground truth and geometric model, as shown by higher mIoU scores. This demonstrates that geometry-aware texture processing improves segmentation, with Texture2LoD3 outperforming manual methods without requiring labor-intensive intervention.

\noindent \textbf{Limitations and Future Work}
The Texture2LoD3 method leverages the worldwide ubiquity of both semantic 3D models and panoramic street-view images, which shall open worldwide availability of so-far scarce \gls{LoD}3 models.
Yet, caution must be exercised as our rectified images are obtained from \gls{GSV} images; there still may be some occlusions present concealing facades; the image quality is also highly dependent on the lighting conditions at the time the \gls{GSV} images were captured.
The identified hyper-parameters were consistently applied to our ReLoD3 dataset, yet further experiments must be undertaken to prove their computational efficiency and scalability, e.g., in architecturally different scenes of Asia.
%
%
\section{Conclusion}
\label{sec:conclusion}
In this paper, we introduce Texture2LoD3, a method enabling \gls{LoD}3 building reconstruction by accurately projecting widely available street-level panoramic images onto surfaces of low-detail semantic 3D building models.
Our work has led us to the conclusion that such a method can unlock worldwide availability of \gls{LoD}3 models, as our automatic results outperform standard projections (by 11\% IoU) and can replace manual texture projections (positive 1\% IoU difference).
Crucially, we also observe the qualitative advantage of our method, as it is less prone to perspective distortions when compared to manual perspective image projection or projecting without any surface rectification.
Moreover, by employing prior low-detail semantic 3D building models as projection targets, we maintain the essential requirements of georeferencing, watertightness, and low-poly representation, extended by texture semantics.
Owing to the absence of datasets allowing for such developments, we present the ReLoD3 texturing benchmark dataset, which will facilitate further research on \gls{LoD}3 building reconstruction from images. 

\noindent {\bf Acknowledgments}
The work was conducted within the framework of the {Leonhard Obermeyer Center} at TUM and the ReLoD3 project of TUM and NUS.
We are grateful for the diligent work of the TUM2TWIN members, especially to Franz Hanke who meticulously collected images and manually projected onto building models.

{
    \small
    \bibliographystyle{ieeenat_fullname}
    \bibliography{main}
}


\clearpage
\setcounter{page}{1}
\maketitlesupplementary

%
\section{Parameter Settings}
\label{sec:parameter_settings}
\noindent \textbf{Processing Hardware}
The experiments were conducted on an OMEN HP Laptop 17 with NVIDIA® GeForce RTX™ 4090 Laptop-GPU (16 GB GDDR6), Intel® Core™ i9-Processor 13. Generation, 32 GB DDR5-5200 MHz RAM (2x 16 GB).
 
\noindent \textbf{B-Rep Preprocessing}
For facade extraction, a ray-casting approach uses multiple rays per camera view. We integrate camera parameters by setting the camera offset to 0.01 m and assuming a camera height of 1.7 m above the building’s lower bound. PCA-based local plane fitting was used for re-triangulation of the fragmented triangular faces.

\noindent \textbf{Geo-Spatial Data Extraction and FOV Computation} 
Building footprints were extracted from CityGML files by parsing the first \texttt{posList} element in the \texttt{GroundSurface}. Coordinates were converted from EPSG:25832 to EPSG:4326. For field-of-view estimation, horizontal angles were interpolated (10 samples) between the adjusted left and right angles, where the inward offset was set as one-twentieth of the overall FOV (e.g., for a 60° FOV, the offset was 3° for both sides). Five pitch samples were also generated within a $\pm5^\circ$ range around the optimal pitch computed from wall surfaces.

\noindent \textbf{Panoramic Image Auto-rectification} 
The rectification module uses default configuration parameters from the original method \cite{9145640}. Each panorama was partitioned into tiles with overlapping regions in our implementation, and a consensus zenith was computed via SVD. The pitch and roll angles for re-projection were derived from the best-fit zenith and further refined by histogram-based aggregation.

\noindent \textbf{Building Facade Segmentation} 
Semantic-SAM was used to generate around 100 to 200 masks on average per street-level image. 
For semantic filtering, a CLIP confidence threshold of 0.05 was applied. 
Subsequent morphological processing used a rectangular kernel from size $25\times25$ to $100\times100$ to ensure the artifacts on the contour's boundary would not influence the quadrilateral fitting; 
we also removed connected components smaller than a certain number of pixels, which was set to 2000 on average.

\noindent \textbf{Facade Mask Quadrilateral Fitting} 
After preprocessing the binary masks with a Gaussian blur (kernel size $25\times25$) and morphological operations, the quadrilateral fitter was applied with the following parameters: Polygons with more than 10 vertices were simplified using an initial epsilon of 0.1, a maximum epsilon of 0.4, and an epsilon increment of 0.02. No additional expansion margin was used. The resulting quadrilaterals were rectified to axis-aligned bounding boxes for perspective transformation.

\noindent \textbf{Texturing by Ray-Casting} Rays were cast from the camera using the 10 interpolated horizontal angles and five pitch samples. Intersection points were projected onto the locally fitted facade plane to compute UV texture coordinates. Texture sampling employs bilinear interpolation to ensure a smooth mapping onto the simplified mesh.

\noindent \textbf{Facade Elements Semantic Segmentation Parameters} 
We utilized the Mask2Former model with a Swin-Large backbone, initializing from weights pre-trained on ADE20K.
We implemented training procedures for both models with consistent hyperparameters: Batch size of four, AdamW optimizer with a learning rate of 5e-5, and weight decay of 1e-4. 
Models were trained for 20 epochs with early stopping based on validation loss. 
Data augmentation included random horizontal flipping and brightness/contrast adjustments to improve generalization.
Evaluation metrics included mean Intersection over Union (mIoU) and per-class IoU.
Visualization of segmentation results alongside ground truth masks provides qualitative insight into model performance, particularly for challenging cases such as closely spaced windows or irregular architectural elements.
Our experimental setup ensured fair comparison across all models by maintaining consistent image resolution, data splits, and evaluation protocols.

\section{Further Details on the Selected Baseline Semantic Segmentation Methods}

We evaluated the performance of four state-of-the-art semantic segmentation approaches on the task of facade opening detection: SegFormer \cite{xie2021segformer}, MaskFormer \cite{cheng2021maskformer}, Mask2Former \cite{cheng2021mask2former}, and Grounded SAM2 \cite{ren2024grounded} (Segment Anything Model with semantic capabilities). 
Each model represents a different architectural paradigm in the evolution of transformer-based segmentation methods. For the close-set supervised methods,
SegFormer \cite{xie2021segformer} combines the hierarchical structure of CNNs with the global modeling capabilities of transformers, utilizing a hierarchical transformer encoder and a lightweight MLP decoder. 
MaskFormer \cite{cheng2021maskformer} approaches semantic segmentation as a mask classification problem rather than per-pixel classification. 
It generates a set of binary masks with associated class predictions, combining the strengths of both semantic and instance segmentation paradigms.
Mask2Former \cite{cheng2021mask2former} advances instance and semantic segmentation through its masked attention mechanism and transformer decoder architecture. For the supervised methods, we leveraged the pre-trained on ADE20K \cite{zhou2017scene}, fine-tuned on the CMP dataset \cite{Tylecek13}. Grounded SAM2 \cite{ren2015faster} extends the capabilities of the Segment Anything Model by incorporating semantic grounding, enabling it to perform semantic segmentation with prompt guidance. For our experiments, we used the text prompt \textit{"window"} and \textit{"door"}.

\section{Geo-Spatial Data Extraction and FOV Computation}
\label{sec:geo_fov}

To complement the model preprocessing, we incorporated a geospatial analysis pipeline that served two purposes: (i) extraction of building footprints in a GIS-friendly format and (ii) computation of the camera’s field-of-view (FOV) for each building.

\noindent \textbf{GeoJSON Conversion from CityGML.} \\
Building models stored in CityGML files were parsed to extract the \texttt{GroundSurface} coordinates. The extracted 3D coordinates (typically in meters) were converted into 2D polygons by retaining the (x,y) components. A coordinate transformation (e.g., from EPSG:25832 to EPSG:4326) was then applied to generate GeoJSON-compliant building footprints. This conversion facilitated integration with external GIS tools and provides a reliable spatial reference for subsequent FOV analysis.

\section{Generation of Cropped Perspective Images with Building ID Labeling}
After determining each panorama's field-of-view (FOV) as described in \cref{sec:geo_fov}, we further generate \emph{cropped perspective images} of the building facades and label them with the corresponding building IDs. The overall pipeline is illustrated on the left side of Figure~\ref {fig:id_association_pipeline}, where each cropped perspective image is annotated with an ID matching the building footprint in the CityGML data.

\paragraph{Overview of the Pipeline}
\begin{enumerate}
    \item \textbf{Panorama Cropping Based on FOV} 
    For each panorama, the relevant horizontal span is identified by computing the left and right boundaries of the view. The panorama is then cropped accordingly to focus on the portion containing the target building facade.

    \item \textbf{Building Region Detection} Detect facade bounding boxes within the cropped panorama using Grounding DINO~\cite{liu2023grounding}, retaining only the highest‑confidence box covering the image center.

    \item \textbf{Perspective Transformation}
    Using the bounding box coordinates, a perspective transformation is applied to extract and rectify the facade. This step accounts for the camera’s heading and pitch, generating a front-to-parallel view of the building surface.

    \item \textbf{Building ID Labeling}
    The resulting perspective image is saved with a filename or metadata embedding the \emph{building ID}. This ID is typically derived from the CityGML data or an external GIS database, ensuring each cropped image can be uniquely matched to the corresponding building footprint.

\end{enumerate}

By following this pipeline, we obtain cropped, perspective-corrected facade images automatically labeled with building IDs. These labeled images are then used to transfer IDs to unlabeled rectified image tiles via feature-based matching (right side of~\cref{fig:id_association_pipeline}). ~\Cref{sec:building_id_association} provides full details of this ID association process.

\section{Building ID Association}
\label{sec:building_id_association}

\begin{figure}[htbp]
    \centering
    \includegraphics[width=1.0\linewidth]{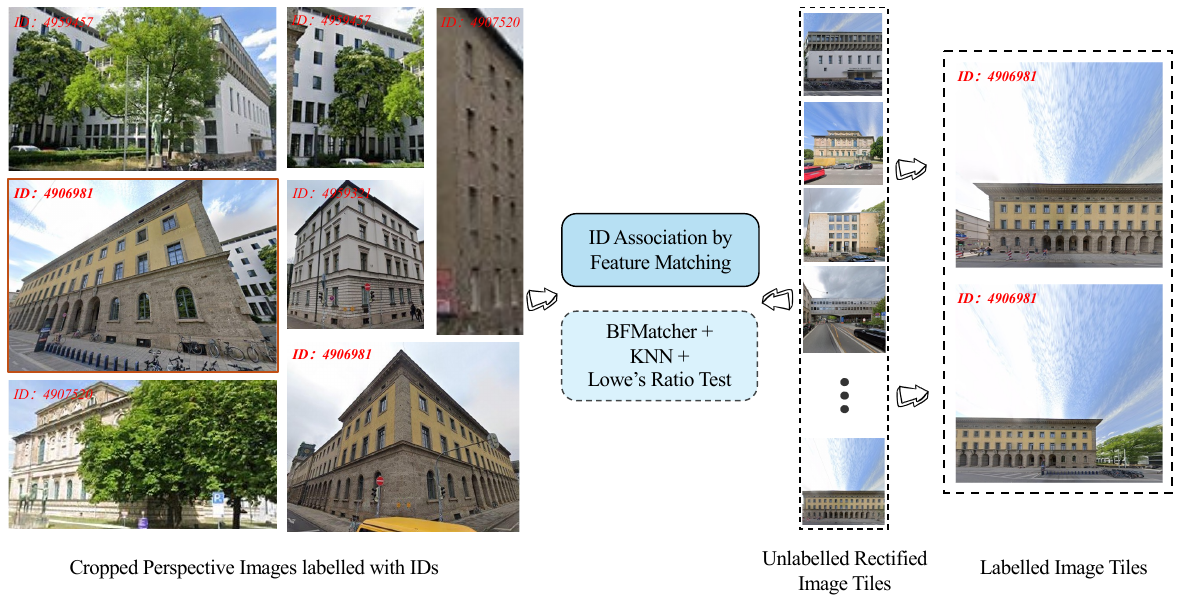} 
    \caption{Pipeline of building ID association. The left side illustrates labeled building images obtained from CityGML data, while the right side presents rectified unlabelled facade tiles. The association is performed using feature matching (BFMatcher + KNN + Lowe’s Ratio Test) to automatically establish correspondences and assign IDs.}
    \label{fig:id_association_pipeline}
\end{figure}

As illustrated in~\cref{fig:id_association_pipeline}, our objective is to automatically associate labeled building images obtained from CityGML data (which contains building \emph{footprints}) with unlabeled rectified image tiles obtained through a generic panorama rectification process. This step enables us to assign building IDs to the previously unlabelled image tiles. The process consists of the following steps:

\begin{enumerate}
    \item \textbf{Data Preparation and Grouping} We begin by extracting unique building IDs from the object detection and CityGML's provided footprints and obtaining labeled building images through projection or rendering processes (left side of ~\cref{fig:id_association_pipeline}). Simultaneously, panorama images are rectified and split into unlabeled tiles that primarily contain building facades and outlines (right side of ~\cref{fig:id_association_pipeline}).

    \item \textbf{Feature Extraction and Matching} To match images of the same building from different perspectives, we employ the SIFT algorithm for keypoint detection and descriptor extraction \cite{lowe2004distinctive}. We further utilize BFMatcher, KNN, and Lowe’s Ratio Test to perform precise feature matching. A threshold on the number of inlier matches is applied to filter out false correspondences.

    \item \textbf{Building ID Association} If a labeled image and an unlabeled rectified tile pass the feature matching threshold (e.g., sufficient inlier matches), we associate the building ID from the labeled image with the rectified tile. This process allows automatic annotation of the previously unlabeled tiles.
\end{enumerate}

By following this approach, the building images with known IDs (examples shown on the left in~\cref{fig:id_association_pipeline}) can be linked with rectified unlabeled facade tiles (examples on the right in~\cref{fig:id_association_pipeline}), enabling automatic ID assignment. Experimental results demonstrate that this method achieves robust and accurate multi-view building matching.

\section{Further Details on the ReLoD3 Texture Dataset Benchmark Creation}
\label{sec:bench_texture}

\noindent \textbf{Extraction of Ground-Truth Openings.} \
We extracted precise opening masks directly from 3D building models in the CityGML format to establish reliable ground truth for evaluating semantic segmentation models on facade openings. 
Our approach leveraged the explicit geometry information available in \gls{LoD}3 building models, where architectural elements such as doors and windows are explicitly modeled. 
The extraction process started by identifying wall surfaces (\texttt{bldg:WallSurface}) in the CityGML file and their associated opening elements. 
For each wall, we extracted the 3D coordinates of the facade polygon and all opening polygons.
These 3D points were then projected onto a 2D plane using Principal Component Analysis (PCA) to obtain the facade's principal plane.
After projection, we converted the 2D points to Shapely \cite{shapely} polygons for geometric operations. 
To address potential topology issues in closely positioned openings (e.g., adjacent windows), we implemented a proximity-based grouping algorithm that merged openings within a specified distance threshold (0.1 meters). 
The facade polygon and opening polygons were combined through boolean operations, where openings were subtracted from the facade to create a comprehensive representation of the wall structure. 
More details are presented under the project page: \textit{[URL anonymized for the submission]}.

\noindent \textbf{Automatic Download of the Street-View Images.}
To efficiently acquire street-view images corresponding to building facades, we have designed an automated download process. This process leverages the implementation of \cite{streetlevel2024}. The workflow is as follows:

\begin{enumerate}
    \item \textbf{Sampling Point Generation} Starting from the predefined start and end coordinates, we use linear interpolation to generate multiple sampling points along the line connecting these coordinates. These points cover the area around the building, ensuring that the collected panorama images contain the relevant building facades.
    \item \textbf{Panorama Query and Download} We query for nearby panorama images for each sampling point. The unique panorama ID is checked against a set of already downloaded IDs to avoid duplicate downloads.
    \item  \textbf{Metadata Recording} During the download process, the script collects metadata for each panorama, including panorama ID, latitude, longitude, heading (in both radians and degrees), capture date, and location; 
    Then, it stores it in a CSV file. This metadata facilitates later association with the CityGML data and further analysis.
\end{enumerate}

\begin{figure}[ht]
    \centering
    \includegraphics[width=0.3\textwidth]{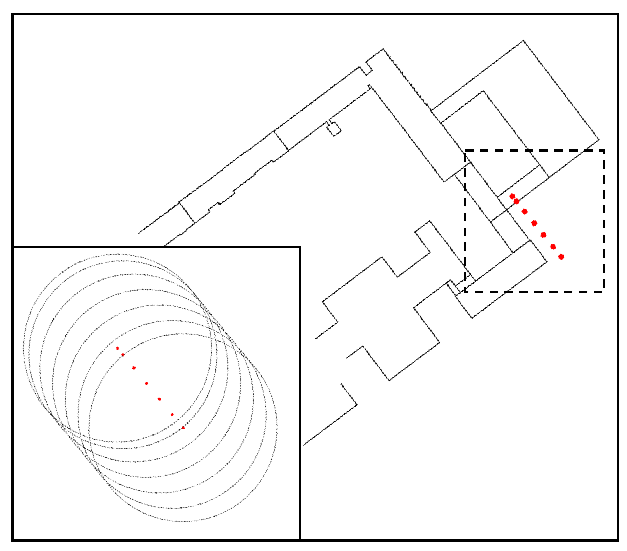}
    \caption{Schematic illustration of building footprint (black), sampling points (red), 
    and the buffer area (gray dashed circles). The buffer defines a maximum distance from each sampling point 
    within which building facades can be captured or considered visible. This ensures coverage of the building's 
    facade from multiple vantage points and avoids unnecessary distant panoramas.}
    \label{fig:buffer_explanation}
\end{figure}

As illustrated in~\cref{fig:buffer_explanation}, the \textit{buffer} is a circular region around each sampling point (with a user-defined radius, e.g., 50 meters). Only those building surfaces (or facade elements) intersecting this buffer are considered relevant for capturing street-view panoramas. This automated workflow ensures high spatial consistency between the street-view images and the building data while significantly improving the efficiency of data collection, thereby providing a robust foundation for subsequent facade texturing and analysis tasks.

\noindent \textbf{Manual 4-point Projection of Perspective Images}
The manually projected perspective terrestrial optical images of the digital camera (Sony $\alpha$7) were acquired specifically for validating automatic texturing purposes.
The campaign was designed to capture the building model facades with a minimum number of photographs per triangle in the existing \gls{LoD}2 building models to ensure texture consistency without any additional image stitching. 

The \textit{4-point projection} refers to the texturing implementation of the proprietary SketchUp Pro \cite{SketchUp} software with the CityEditor \cite{CityEditor} plugin.
While the default SketchUp Pro allows for the manual identification of four image-to-model projection points, the CityEditor allows the loading of CityGML building models into the SketchUp software.
Additionally, \gls{LoD}3 ground-truth models were loaded to guide the manual projection process.
Nevertheless, owing to still persistent distortions, the deviations between the ground-truth \gls{LoD}3 and manual projection exist. 
As such, the distortion-free and cm-accurate \gls{LoD}3 masks shall be treated as the ground truth.

\section{Texturing after triangulation}
\label{sec:texturing_after_triangulation}
\begin{figure}[t]
    \centering
    \includegraphics[width=1.0\linewidth]{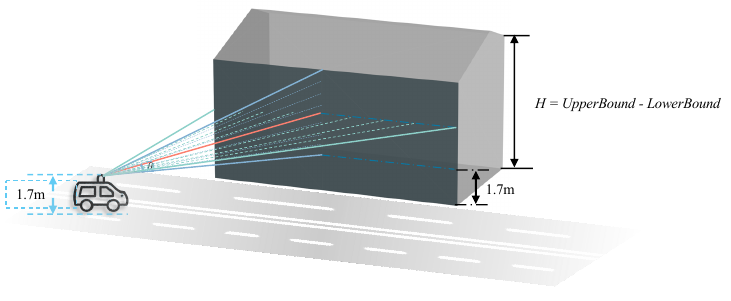}
    \caption{An illustration of our raycasting-based texturing setup. The camera (e.g., mounted on a vehicle at 1.7\,m height) casts multiple rays toward the building's facade, which extends from the lower bound to the upper bound obtained from the GML data. We sample horizontal angles between the left and right viewing directions and interpolate a small range of pitch angles to capture the relevant parts of the facade.}
    \label{fig:rays_estimation}
\end{figure}

We first employ our wireframe preprocessing pipeline (\cref{sec:wireframe_preprocessing}) to enable robust texturing of building facades to convert highly subdivided B-Reps into minimal quadrilateral faces. After this simplification step, we perform ray casting from known camera poses to identify which faces are visible from each viewpoint. \Cref{fig:rays_estimation} illustrates how the camera, positioned at 1.7\ m above the ground, casts rays spanning a specified field of view. The building facade's lower and upper vertical bounds are derived from CityGML data, ensuring that our texturing pipeline only samples the relevant portions of the geometry. For each B-Rep:
\begin{enumerate}
    \item We compute the camera origin and direction based on geographic coordinates and a small offset from the facade.
    \item We cast multiple rays spanning the horizontal viewing angles (from left to right and a range of pitch angles around the facade’s center.
    \item We collect all intersected faces and compute appropriate UV coordinates for texturing. Faces whose normals point inwards are automatically flipped to ensure the texture is placed on the exterior surface.
\end{enumerate}

Finally, once all relevant faces are identified, we project the corresponding panoramic images onto these faces using a planar mapping approach (\cref{alg:texturing_code}). 
This step ensures that the final textured facade remains visually coherent and avoids the distortions that can arise when projecting onto densely triangulated B-Reps. 
The resulting textured model forms the basis for subsequent facade analysis and segmentation (\cref{sec:facade_seg}).

\section{Building Facade Segmentation: Influence of Candidate Masks}

This step aims to detect and isolate the main building facade from the textured geometry. Our approach employs a semantic segmentation pipeline built upon Semantic-SAM, which automatically generates a set of candidate masks for each panoramic or perspective image. We then filter these masks to retain only those corresponding to the "building facade'' class, discarding irrelevant classes such as sky, road, or cars. Small floating artifacts are removed via connected-component analysis, and we apply morphological smoothing to obtain a clean, consolidated facade mask suitable for further processing.

\begin{figure}[t]
    \centering
    \begin{subfigure}[t]{0.48\linewidth}
        \centering
        \includegraphics[width=\linewidth, trim=0 10 0 0, clip]{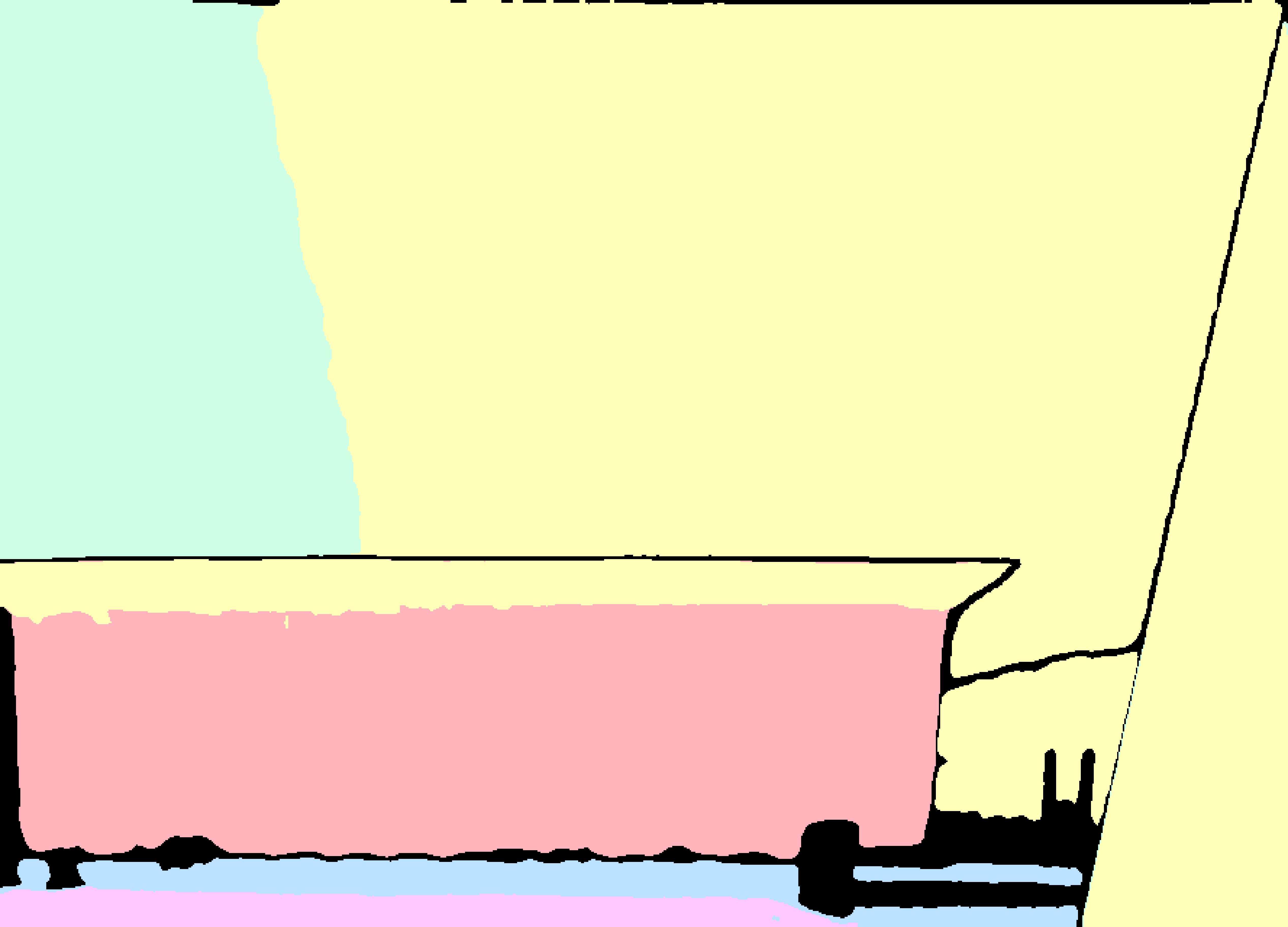}
        \caption{Coarse segmentation (10 masks)}
        \label{fig:seg_coarse}
    \end{subfigure}
    \hfill
    \begin{subfigure}[t]{0.48\linewidth}
        \centering
        \includegraphics[width=\linewidth, trim=0 10 0 0, clip]{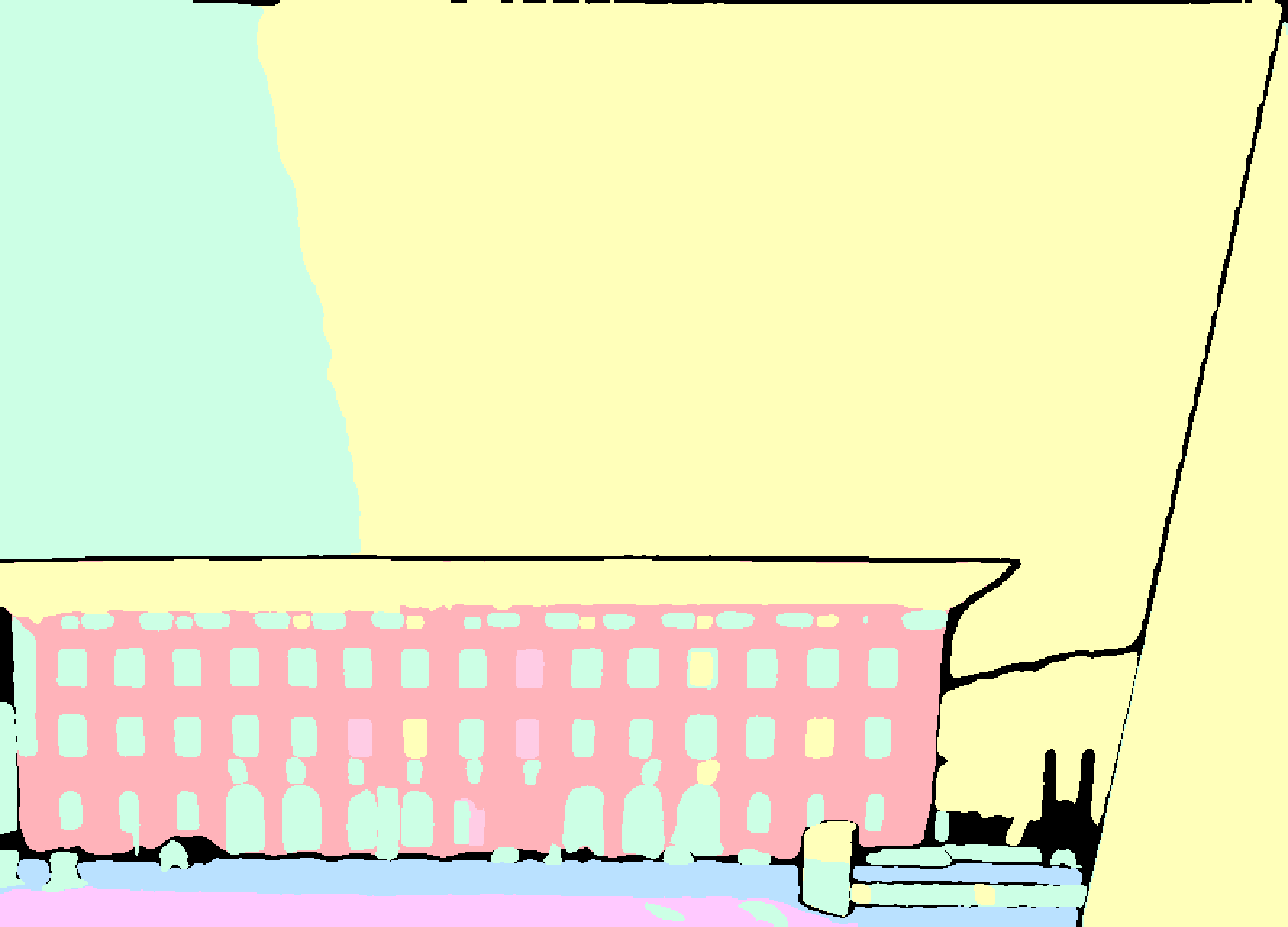}
        \caption{Fine‑grained segmentation (127 masks)}
        \label{fig:seg_fine}
    \end{subfigure}
    \caption{Comparison of segmentation results using different numbers of retained candidate masks. A small number of masks (left) leads to fewer, larger segments capturing the main facade region. In contrast, a larger number of masks (right) produces more detailed but also more fragmented subregions.}
    \label{fig:topk_comparison}
\end{figure}

~\Cref{fig:topk_comparison} demonstrates how adjusting the quantity of retained candidate masks affects the final segmentation. In ~\cref{fig:topk_comparison}(a), retaining only 10 masks results in coarser segmentation with fewer, larger regions that effectively capture the overall facade shape. Such coarse segmentation is often advantageous when the primary goal is to isolate the facade with minimal clutter. Conversely, ~\cref{fig:topk_comparison}(b) shows a more fine-grained segmentation derived from 127 candidate masks, revealing additional details such as windows or ornamental features. While this can benefit downstream tasks requiring higher granularity, it also increases the likelihood of fragmented subregions that complicate facade isolation.
%
%
%
%

\begin{figure*}[ht]
    \centering
    \includegraphics[width=\linewidth]{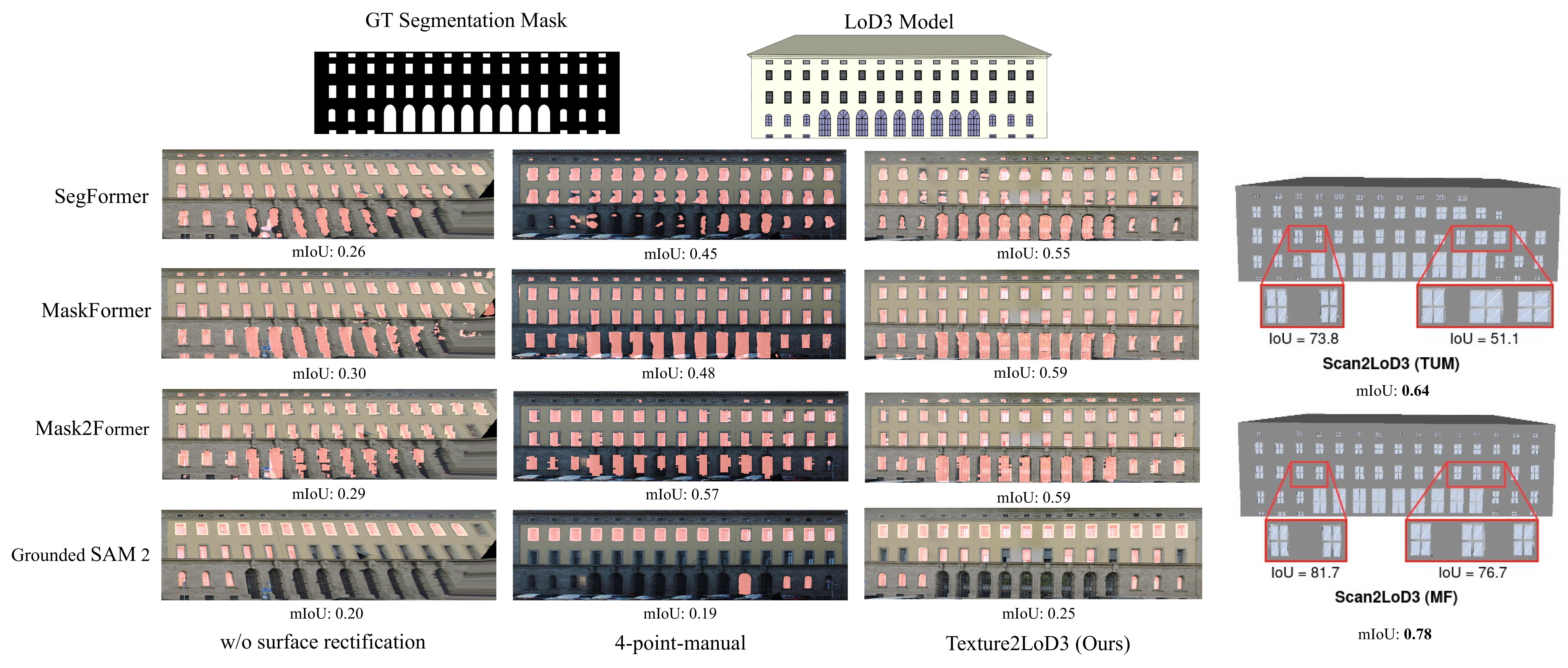}
    \caption{Comparison of the baselines and the Texture2LoD3 method to the Scan2LoD3 method leveraging multi-modal fusion of laser scanning, 3D model priors, and street-level images. Such an approach clearly outperforms only image and model combinations. Yet such a multi-modal setup is scarcely available in practical scenarios, unlike street-level images and 3D models. Figure parts copied and edited from the original Scan2LoD3 article, where experiments were conducted on the same object, courtesy of \citet{wysocki2023scan2lod3}.}
    \label{fig:scan2lod3_compare}
\end{figure*}
\section{Test-time Alignment for Mask Evaluation}
Due to the inherent transformation challenges in panorama rectification, we implement a test-time scale and shift adjustment procedure when evaluating predicted segmentation masks against ground truth masks. This adjustment is necessary because the rectification process introduces unavoidable geometric distortions, causing the segmented objects to lose their absolute scale and position relative to the original panoramic view. Our method employs a two-stage optimization approach: First, conducting a coarse grid search over a constrained parameter space (scale factors between 0.75 and 1.2, and pixel translations within $\pm$100 pixels), followed by a finer search within a more focused range around the best parameters identified in the first stage. For each candidate transformation, we compute the Intersection over Union (IoU) between the predicted mask and the transformed ground truth mask, selecting the parameters that maximize this metric. This alignment procedure ensures a fair comparison between prediction and ground truth by compensating for the scale and positional discrepancies introduced during the rectification process without altering the structural integrity of the segmentation boundaries.


\section{Comparison to the Scan2LoD3 method}
As mentioned in Related Work (Section \ref{sec:rw}), there are methods leveraging the accuracy of laser scanning, building priors, and images to reconstruct LoD3 building models.
We acknowledge that this approach yields superior performance to our work owing to the use of accurate laser scanning modality and physics-oriented ray analysis.
Due to that fact, this comparison is out of the scope of the main publication part. 
Nevertheless, such a comparison is worth showcasing modalities' limitations, primarily since experiments were performed partially on the same objects. 
Here, we selected an excerpt from the \citet{wysocki2023scan2lod3} Scan2LoD3 method that performed the analysis on the same building (the so-called \textit{building 23}). As we show in \Cref{fig:scan2lod3_compare}, the performance on the same facade increases significantly owing to the laser scanner modality. It scored 78\% while using high accuracy scanner, and 64\% when using lower grade Velodyne scanner. 
This experiment shows a minimum of 5\% and a maximum of 14\%  increase compared to the best baseline image-based segmentation.
Yet, as we elaborate in Related Work (Section \ref{sec:rw}), such a multi-modal setup is still scarcely available, in contrast to the ubiquitous street-level images and 3D prior models.

\end{document}